\documentclass[letterpaper, 10 pt, conference]{ieeeconf}  

\IEEEoverridecommandlockouts                              

\usepackage{amsmath}

\usepackage{amssymb}
\usepackage{graphicx}
\usepackage{caption}
\usepackage{booktabs}
\let\labelindent\relax
\usepackage[shortlabels]{enumitem}
\usepackage{twemojis}
\usepackage{tabularx}
\usepackage{multirow}
\usepackage{url}
\usepackage{subcaption}
\usepackage{quoting,xparse}
\usepackage{kotex}
\usepackage{comment}
\usepackage{eso-pic}

\title{\LARGE \bf
AdaVLA: Adaptive Step Flow Matching for Training-free Acceleration of Vision-Language-Action Models
}

\author{Sunghwan Han$^{1}$, Youngtae Han$^{1}$, and Youngmin Yi$^{1\dagger}$
\thanks{$^{1}$All authors are with the Dept. of Artificial Intelligence, Sogang University, Seoul, 04107, Republic of Korea. \{\tt\footnotesize sunghwan06, young0tete, ymyi\}@sogang.ac.kr}
\thanks{$\dagger$ Corresponding author.}
}

\newcommand{\SYS}{\textsl{AdaVLA}}
\newcommand{\ymyif}[1]{{\color{black} #1}}
\newcommand{\ymyi}[1]{{\color{black} #1}}
\newcommand{\hesh}[1]{{\color{black} #1}}
\newcommand{\camera}[1]{{\color{black} #1}}

\begin{document}

\AddToShipoutPictureFG*{%
  \AtPageLowerLeft{%
    \raisebox{6mm}{%
      \makebox[\paperwidth][c]{%
        \parbox{0.92\textwidth}{%
          \centering
          \fontsize{6}{6.8}\selectfont
          \copyright~2026 IEEE.
          Personal use of this material is permitted.
          Permission from IEEE must be obtained for all other uses,
          in any current or future media, including reprinting/republishing
          this material for advertising or promotional purposes,
          creating new collective works, for resale or redistribution
          to servers or lists, or reuse of any copyrighted component
          of this work in other works.
        }%
      }%
    }%
  }%
}

\maketitle
\thispagestyle{empty}
\pagestyle{empty}

\begin{abstract}
Vision-Language-Action (VLA) models, built upon Vision-Language Models (VLMs), have significantly enhanced robotic capabilities by leveraging internet-scale knowledge and \hesh{multimodal} reasoning. However, the \ymyi{intensive} computational overhead of VLAs \ymyi{constrains on-device} deployment, hindering \ymyi{real-time} responses to \ymyi{environmental} changes.

While various acceleration techniques have been proposed, they often rely on fine-tuning or access to training datasets, which are frequently unavailable due to privacy and proprietary concerns. 
Moreover, although flow-matching-based VLAs
have emerged as efficient alternatives to standard diffusion models, current acceleration efforts largely target VLM inference costs, failing to address the
\hesh{iterative ODE solving process inherent in flow matching inference.}
To address these limitations, we propose \SYS, an online, training-free \ymyi{adaptive framework for fast yet accurate} flow-matching-based Vision-Language-Action models. We introduce a novel metric derived from the \hesh{flow matching trajectory curvature} to quantify action generation confidence during inference. This metric enables the dynamic reduction of inference steps and the adaptive adjustment of MLP pruning ratios through an efficiently \hesh{computed} importance \ymyi{evaluation}, requiring no access to training data.
Experimental results on the LIBERO benchmark using a Jetson \ymyi{AGX} Orin device demonstrate that our method achieves $1.87\times$ and $2.24\times$ speedups for $\pi_{0.5}$ and X-VLA, respectively, with negligible degradation in success rates. Furthermore, we validate the robustness of our approach on real-world robotic tasks using SmolVLA.

\end{abstract}

\section{INTRODUCTION}
Recent advancements in integrating visual and language models have led to the development of Vision-Language Models (VLMs), marking significant progress in multimodal reasoning and generalization. This progress has extended to robotic action planning, evolving VLMs from generating textual responses to \hesh{predicting} direct control actions, resulting in Vision-Language-Action (VLA) models~\cite{kim2024openvla,brohan2023rt2,octo_2023}. These models demonstrate superior robustness and generalization across diverse environments compared to traditional reinforcement learning policies.

However, VLAs inherit the \ymyi{intensive} computational costs of the underlying VLM. This results in substantial inference latency between actions, leading to 
unnatural robotic motion. While several acceleration methods—including
\hesh{lightweight design} (TinyVLA~\cite{wen2025tinyvla}), early-exiting (DeeR-VLA~\cite{yue2024deervla}), and \hesh{token} pruning with layer-skipping (EfficientVLA~\cite{yang2025efficientvla})—
have been proposed, they typically require access to training data for fine-tuning or calibration. 
\ymyi{In many practical settings, however, such access is restricted due to proprietary concerns, and the sheer volume of these datasets often prohibits processing on 
edge devices.}

Furthermore, current acceleration strategies~\cite{zhang2025sparsevlm,chen2024image} primarily focus on reducing the computational load of the VLM backbone. However, the emergence of flow-matching~\cite{lipman2023flow}-based VLAs, such as the $\pi$ series~\cite{black2024pi0visionlanguageactionflowmodel,intelligence2025pi05visionlanguageactionmodelopenworld}, SmolVLA~\cite{shukor2025smolvla}, and X-VLA~\cite{zheng2026xvla}, has changed this landscape. In these architectures, the VLM serves as an encoder that passes context via \hesh{the key–value (KV) cache} to the flow-matching-based Action Expert through cross-attention. As illustrated in our subsequent latency analysis (Sec.~\ref{pre:latency}), the iterative process of generating actions from noise within the Action Expert now accounts for a larger portion of the total latency than the VLM backbone. Consequently, optimizing only the VLM is insufficient for accelerating state-of-the-art flow matching VLAs. 

\begin{figure}[tbp]
    \centering
    \includegraphics[width=\columnwidth]{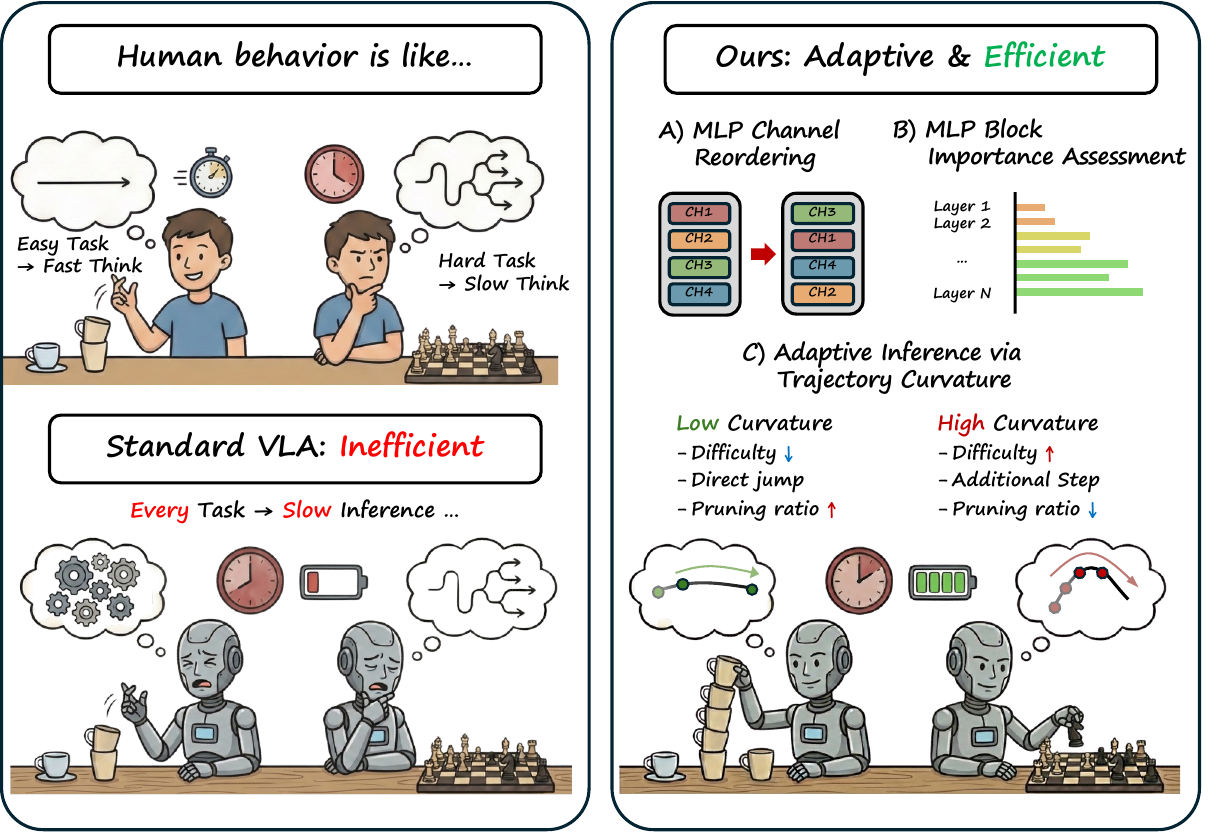} 
    \caption{Concept of \hesh{\SYS}, our proposed online adaptive acceleration framework.}
    \vspace{-20pt}
    \label{fig:paper_concept}
\end{figure}

To address these challenges, we propose \hesh{\SYS}, a training-free, online adaptation framework that \ymyi{simultaneously} reduces computational costs for both the VLM and the Action Expert during runtime. 
Drawing inspiration from human cognition—\textit{where simple tasks elicit quick, intuitive responses while complex ones demand deeper deliberation}—we enable flow-matching-based VLAs to autonomously \ymyi{adapt} their computational effort based on the perceived difficulty of the task at hand, as illustrated in Fig.~\ref{fig:paper_concept}. 

Our methodology \hesh{consists of} three key components that serve as an end-to-end pipeline for acceleration: 
\begin{enumerate}[(1)]
    \item \textbf{Curvature-based Adaptive Inference}: We \ymyi{leverage} the \hesh{flow-matching trajectory curvature to quantify action generation confidence during inference}
    and dynamically \ymyi{adapt} both the inference step size and the \ymyi{MLP} pruning ratio during the \hesh{ODE solving} process (Sec.~\ref{subsec:adaptive_inference}).
    \item \textbf{MLP Block Importance Assessment}: \ymyi{Since naively pruning MLP channels across all layers in a uniform manner can degrade the success rate,} we introduce an efficient, SVD-free assessment to evaluate the importance of an entire MLP block \ymyi{in each layer}. This allows the framework to \ymyi{efficiently} preserve representational diversity while \ymyi{simultaneously} reducing computational costs (Sec.~\ref{subsec:mlp_importance}).
    \item \textbf{MLP Channel Reordering}: We reorder intermediate channels \ymyi{within an MLP block} in descending order of an importance metric \ymyi{to account for dynamic changes in importance. This is selectively triggered} during the initial forward pass or upon detecting a significant context shift, enabling \ymyi{efficient yet effective} structured pruning without training data access (Sec.~\ref{subsec:mlp_reordering}).
\end{enumerate}
Through evaluations on $\pi_{0.5}$~\cite{intelligence2025pi05visionlanguageactionmodelopenworld} and X-VLA~\cite{zheng2026xvla} using a Jetson AGX Orin, we demonstrate 1.87$\times$ to 2.24$\times$ latency \hesh{reductions} on the LIBERO benchmark~\cite{liu2023libero} with minimal impact on the success rate. We also demonstrate the effectiveness \hesh{of \SYS} in real-world robotic tasks on the SO-ARM101~\cite{knight2024standardopen} using SmolVLA~\cite{shukor2025smolvla}.

\section{RELATED WORK}

\subsection{Vision-Language-Action (VLA) Models}
VLA models have emerged as a new paradigm for general-purpose robot learning. Early developments such as RT-1~\cite{brohan2022rt1} utilized Transformers to output discrete actions, while RT-2~\cite{brohan2023rt2} demonstrated that co-fine-tuning VLMs on large-scale datasets
enables the transfer of semantic reasoning to robotic control tasks. Moving beyond autoregressive policies like OpenVLA~\cite{kim2024openvla}, continuous generative approaches have been introduced \hesh{for modeling} complex action distributions. Diffusion Policy~\cite{chi2023diffusion} pioneered the application of diffusion models to visuomotor control, leading to integrated VLA architectures such as CogACT~\cite{li2024cogact} and DiffusionVLA~\cite{wen2025diffusionvla}. Recently, flow matching~\cite{lipman2023flow} models, including $\pi_0$~\cite{black2024pi0visionlanguageactionflowmodel}, $\pi_{0.5}$~\cite{intelligence2025pi05visionlanguageactionmodelopenworld}, SmolVLA~\cite{shukor2025smolvla}, and X-VLA~\cite{zheng2026xvla}, have gained prominence. These models leverage continuous normalizing flows to achieve high-quality generation with significantly fewer steps, making them effective for real-time control.

\subsection{Efficient Vision-Language-Action Models}
Despite their capabilities, the substantial computational costs of VLAs hinder their deployment on resource-constrained edge devices. To address this \ymyi{challenge}, various acceleration techniques have been proposed. RoboMamba~\cite{liu2024robomamba} introduced State Space Models (SSMs)~\cite{gu2024mamba} for \ymyi{robotic} reasoning \ymyi{tasks}, \ymyi{achieving} linear complexity with respect to sequence length. DeeR-VLA~\cite{yue2024deervla} proposed a dynamic early-exiting mechanism, allowing the model to terminate inference early for simple tasks to reduce computation. OpenVLA-OFT~\cite{kim2025finetuning} enables efficient parallel decoding for large-scale backbones. Other efforts mitigate Transformer redundancy; for instance, EfficientVLA~\cite{yang2025efficientvla} and MoLe-VLA~\cite{zhang2025mole} utilize token/layer pruning and dynamic layer-skipping to enhance efficiency.
\hesh{However,}
\ymyi{these approaches rely on retraining or calibration (e.g., layer importance assessment), both of which \hesh{typically} require training data. This dependency limits their applicability in data-limited and privacy-sensitive scenarios, which our work addresses.}

\section{Preliminary}

\subsection{Formulation and VLM Encoding}
A Vision-Language-Action (VLA) model learns a policy $p(A_t | o_t)$ that maps an observation $o_t$ at time $t$ to an action chunk $A_t \in \mathbb{R}^{H \times D}$, where $H$ denotes the action horizon and $D$ represents the action dimension. The observation $o_t$ is a multimodal input consisting of images $I_t^1, \dots, I_t^n$, a language instruction $\ell_t$, and the current robot state $q_t$:
\begin{equation}
    o_t = [I_t^1, \dots, I_t^n, \ell_t, q_t]
\end{equation}
The VLM encodes this observation $o_t$ through a single forward pass. During this process, the resulting Key-Value (KV) cache is shared with the Action Expert to perform cross-attention, allowing the model to utilize multimodal information while generating actions from noise.

\subsection{Action Expert \& Flow Matching}\label{pre:flow_matching}
The Action Expert generates the action $A_t$ using a conditional flow matching~\cite{lipman2023flow} framework \hesh{that learns a vector field along an \hesh{interpolation} path from noise to data.} A linear path between the Gaussian noise $\epsilon$ and the clean action $A_t$, parameterized by the flow-matching time variable $\tau \in [0,1]$, is defined as follows:
\begin{equation}
    A_t^\tau = \tau A_t + (1 - \tau)\epsilon, \quad \text{where } \epsilon \sim \mathcal{N}(\mathbf{0}, \mathbf{I})
\end{equation}
During training, the model learns \hesh{a vector field that induces a trajectory from the noise to the clean action. The target vector field $\mathbf{u}$ along this path is given by the derivative of $A_t^\tau$ with respect to $\tau$:} 
\begin{equation}
    \mathbf{u}(A_t^\tau | A_t) = \frac{d}{d\tau} A_t^\tau = A_t - \epsilon
\end{equation}
Since the target vector field $\mathbf{u}$ is $\tau$-independent, the induced target trajectory is linear. This formulation enables efficient ODE integration and can reduce the number of solver steps in practice.
The model $\mathbf{v_\theta}$ is trained to approximate the target vector field $\mathbf{u}$ by minimizing the squared error between them, with $\tau$ uniformly sampled:
\begin{equation}
    \mathcal{L}(\theta) = \mathbb{E}_{\;\tau, \epsilon, (o_t, A_t)} \left[ \left\| \mathbf{v_\theta}(\tau, A_t^\tau, o_t) - \mathbf{u}(A_t^\tau | A_t) \right\|^2 \right]
\end{equation}
At inference, starting from the noise state ($\tau = 0$), the model solves the Ordinary Differential Equation (ODE) toward the data state ($\tau = 1$) \hesh{conditioned on the observation $o_t$.
Specifically, an Euler solver is employed to 
transform an initial noise sample into the action sequence.}
\begin{table}[t]
\centering
\caption{
\ymyi{Inference latency of $\pi_{0.5}$ on Jetson AGX Orin}
}
\label{tab:latency}
\begin{tabular*}{\columnwidth}{l @{\extracolsep{\fill}} l l}
\toprule
\textbf{Module} & \textbf{Component} & \textbf{Latency (ms)} \\ \midrule
\textbf{VLM Backbone} & Processing Input & 96.25 \\
(18 Layers) & VLM Forward Pass & 290.28 \\
&  $\llcorner$ Self-Attention &  $\llcorner$ 3.05 \\
&  $\llcorner$ MLP Block &  $\llcorner$ \textbf{12.47} \\ \midrule
\textbf{Action Expert} & Flow Matching Process & \textbf{557.78} \\
(10 steps) &  $\llcorner$ Single Step & $\llcorner$ 53.75 \\ \midrule
\textbf{Total Inference} & End-to-End & 953.62 \\ \bottomrule
\end{tabular*}
\vspace{-15pt}
\end{table}

\begin{figure*}[t]
    \centering
    \includegraphics[width=0.93\textwidth]{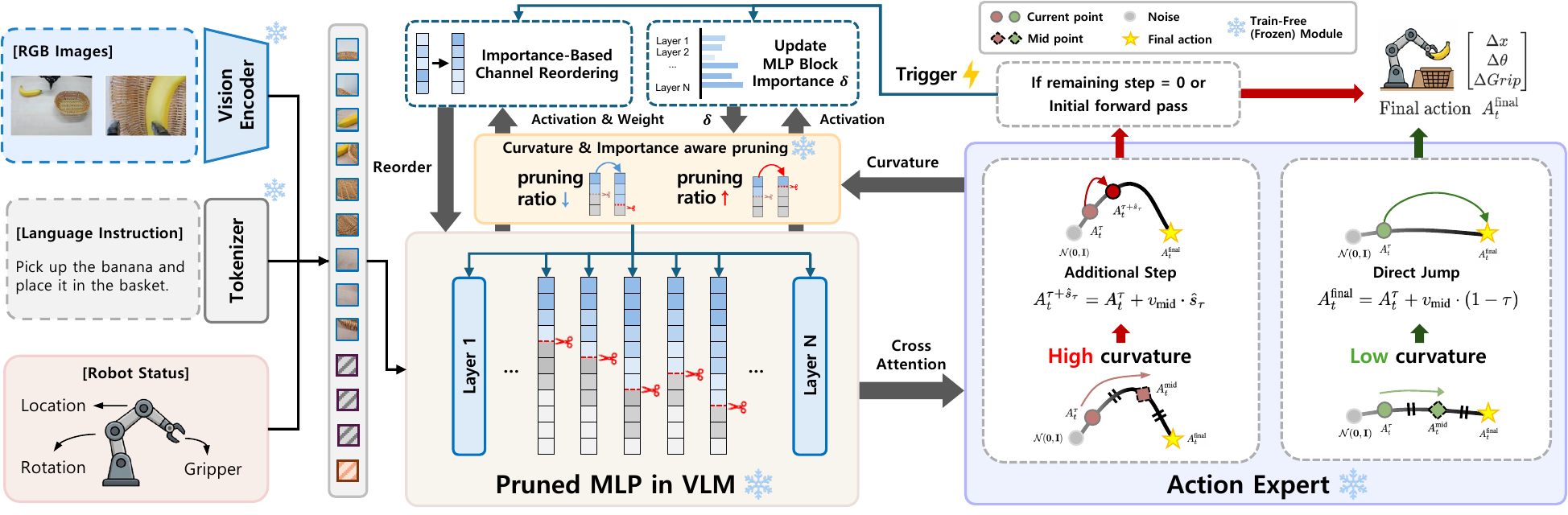} 
    \caption{Overall architecture of \hesh{\SYS}, our proposed online adaptive acceleration framework.}
    \label{fig:method_overview}
    \vspace{-15pt}
\end{figure*}
\subsection{Latency Analysis}\label{pre:latency}
To analyze the \ymyi{performance} bottlenecks on \ymyi{edge devices},
we measured the average inference latency of $\pi_{0.5}$~\cite{intelligence2025pi05visionlanguageactionmodelopenworld} on a Jetson AGX Orin while performing the LIBERO benchmark~\cite{liu2023libero}. The latency profile for action generation is summarized in Table~\ref{tab:latency}, and the key findings from the analysis are as follows:

\begin{itemize}[leftmargin=*]
    \item \textbf{VLM Bottleneck (MLP vs. Attention):}
    \hesh{The VLM backbone} accounts for approximately 40\% (386.53 ms) of the total inference latency. Within its 18 layers, the MLP blocks (12.47 ms per layer) are significantly more computationally intensive than the self-attention blocks (3.05 ms per layer), which aligns with recent findings in LLM optimization~\cite{lee2024cats, wei2024building}.
    \hesh{Although the Action Expert consumes a larger proportion of the latency, targeting MLP blocks within the VLM provides an essential control knob to optimally balance the latency--success rate trade-off.}
    \item \textbf{Action Expert Bottleneck (Iterative ODE Solving):} The flow-matching process constitutes the primary bottleneck, representing approximately 58\% (557.78 ms) of the total inference time. While individual steps are relatively fast (53.75 ms), the cumulative latency of 10 iterations is substantial. 
    \ymyi{This highlights that adjusting the number of iterations based on trajectory curvature is essential for improving overall efficiency.}
\end{itemize}

\section{Method}

Fig.~\ref{fig:method_overview} illustrates the overall workflow of the proposed framework, \SYS. Our methodology comprises three primary components: (i) \textbf{Adaptive Inference} (Sec.~\ref{subsec:adaptive_inference}), which reduces redundant ODE solver steps during action generation within the Action Expert; 
(ii) \textbf{MLP Block Importance Assessment} (Sec.~\ref{subsec:mlp_importance}), \ymyi{a training-free} and efficient scheme to modulate VLM backbone pruning ratios by leveraging Participation Ratio-based block importance;
and 
(iii) \textbf{MLP Reordering} (Sec.~\ref{subsec:mlp_reordering}), which facilitates structured pruning by accounting for channel-wise importance.
 
\subsection{Adaptive Inference via Trajectory Curvature}\label{subsec:adaptive_inference}

Generating actions via flow matching is a primary bottleneck due to its iterative nature. We propose an online adaptation framework that dynamically adjusts \ymyi{flow matching} step sizes and pruning ratios based on perceived difficulty. 

\hesh{As shown in Sec.~\ref{pre:flow_matching}, the flow matching model learns a vector field that transports samples from the noise $\epsilon$ toward the action $A_t$.} Unlike high-dimensional image generation, where straight trajectories are rarely guaranteed due to immense dimensionality,
action generation operates in a significantly more compact space.
Given that these action dimensions are substantially smaller than \hesh{those of} high-dimensional image spaces, we hypothesize that
\hesh{the induced sample trajectory would remain nearly linear unless the model encounters an uncertain state.}
Consequently, the straightness of trajectories can serve as a reliable proxy for model confidence. To quantify trajectory deviation, we define the relative curvature $C$ between \hesh{estimated vectors} $v_{\tau_A}$ and $v_{\tau_B}$ ($\tau_A < \tau_B$), using the magnitude of $v_{\tau_B}$ as a normalization factor, where $\varepsilon$ is a small constant for numerical stability:
\begin{equation}
    C = \frac{\|v_{\tau_A} - v_{\tau_B}\|_2}{\|v_{\tau_B}\|_2 + \varepsilon}
\end{equation}
Unlike standard flow matching's Euler steps, \SYS{} draws inspiration from the second-order Runge-Kutta (RK2) method~\cite{runge1895numerische, kutta1901beitrag} not only for \ymyi{enhanced} trajectory estimation but \ymyi{also as an indicator of trajectory straightness.} 
Although RK2 doubles the number of function evaluations (NFE = 2) per iteration, \SYS{} strategically bypasses redundant computations when the trajectory is nearly linear.
At time $\tau$, \SYS{} computes $v_\tau := \mathbf{v_\theta}(\tau, A_t^\tau, o_t)$, sets $\tau_{\text{mid}}=\tau+\frac{1-\tau}{2}$ and $A_t^{\text{mid}}=A_t^\tau+v_\tau\cdot\frac{1-\tau}{2}$, and then evaluates $v_{\text{mid}}:=\mathbf{v_\theta}(\tau_{\text{mid}},A_t^{\text{mid}},o_t)$.
Then, the curvature $C$, computed from $v_\tau$ and $v_{\text{mid}}$, is compared against a predefined threshold $C_{\text{th}}$, based on which the inference process branches dynamically.

\textbf{Low Curvature ($C < C_{\text{th}}$):} The trajectory is interpreted as nearly linear. 
Relying on $v_{\text{mid}}$ to extrapolate the trajectory, the framework bypasses all remaining iterations to execute a \textit{direct jump} to the final state ($\tau\approx1$):
\begin{equation}
    A_t^{\text{final}} = A_t^\tau + v_{\text{mid}} \cdot (1 - \tau)
\end{equation}
Simultaneously, we aggressively increase the $l$-th layer's base pruning ratio $p_l$, modulated by the layer importance $\delta_l$, while ensuring \ymyi{that} it remains within a valid range:
\begin{equation}
    p_l' = \min \left( 1.0,\;p_l + \exp(1 - \delta_l) \cdot (C_{\text{th}} - C) \right)
\end{equation}
\ymyi{where $\delta_l$ and $p_l$ are detailed in Sec.~\ref{subsec:mlp_importance}.} 

\textbf{High Curvature ($C \ge C_{\text{th}}$):} The increased trajectory complexity necessitates a more conservative yet adaptive progression. Leveraging the pre-computed midpoint vector $v_{\text{mid}}$, we advance the state using a bounded step size $\hat{s}_\tau$. Since the RK2 method requires two function evaluations (NFE = 2) per iteration, we aim to utilize a step size of $2s$ to maintain computational efficiency comparable to standard solvers. However, to prevent success rate degradation in regions of high complexity, we modulate this target step size by $C_{\text{th}}/C$, allowing the model to dynamically scale down its progression based on the measured \hesh{trajectory} deviation. \hesh{To ensure stability, we bound $\hat{s}_\tau$ to the range $[s,\,(1-\tau)/2]$:}
\begin{equation}
    \hat{s}_\tau = \min \left( \frac{1-\tau}{2}, \max \left( s, \frac{C_{\text{th}}}{C} \cdot 2s \right) \right)
\end{equation}
\hesh{
We use $v_{\text{mid}}$ for updates because, compared to the current-time evaluation $v_\tau$, it is empirically more reliable (lower-variance) and yields more stable updates under the same compute budget.}
The next state is then updated as follows:
\begin{equation}
    A_t^{\tau + \hat{s}_\tau} = A_t^\tau + v_{\text{mid}} \cdot \hat{s}_\tau
\end{equation}
To restore reasoning capacity, the pruning ratio is decreased 
based on the layer importance $\delta_l$:
\begin{equation}
    p_l' = \max \left( 0,\; p_l - \exp(\delta_l) \cdot (C - C_{\text{th}}) \right)
\end{equation}
If the maximum allowed steps are exhausted, \hesh{\SYS} interprets the situation as a significant context shift, triggering the recalculation of MLP block importance (Sec. \ref{subsec:mlp_importance}) and MLP channel reordering (Sec. \ref{subsec:mlp_reordering}).

\hesh{By consolidating these adaptive functionalities, \SYS{} encapsulates the latency--success rate trade-off into a single \ymyi{design} parameter, the curvature threshold $C_{\text{th}}$.
This eliminates the need for manual, multi-dimensional tuning of layer-wise pruning ratios and flow matching hyperparameters.
As shown in the ablation study (Sec.~\ref{exp:ablation}), there is a gradual trade-off between success rate and latency as $C_\text{th}$ increases, allowing for scenario-specific optimization.}
\begin{table*}[t]
\centering
\caption{\ymyi{Comparison} of \SYS{} with $\pi_{0.5}$ and X-VLA on the LIBERO Benchmark. }
\label{tab:libero_main}
\begin{tabular}{@{}lclcccccccc@{}}
\toprule
\multirow{2}{*}{\textbf{Base Model}} & \multirow{2}{*}{\textbf{Params}} & \multirow{2}{*}{\textbf{Method}} & \multicolumn{5}{c}{\textbf{Success Rate (\%)}} & \multicolumn{2}{c}{\textbf{Latency (ms)}} & \textbf{Energy (J)} \\ \cmidrule(lr){4-8} \cmidrule(lr){9-10} \cmidrule(l){11-11} 
 & & & Spatial & Object & Goal & Long & \textbf{Average} & A10 & {Orin} & {Orin} \\ \midrule
\multirow{4}{*}{$\pi_{0.5}$~\cite{intelligence2025pi05visionlanguageactionmodelopenworld}} 
 & \multirow{4}{*}{3.3B} & Baseline & \textbf{99.40} & \underline{98.20} & \underline{98.80} & \underline{98.20} & \underline{98.65} & 289.68 & 958.54 & 19.83 \\
 & & + SparseVLM~\cite{zhang2025sparsevlm} & \underline{96.60} & 97.00 & 97.60 & 94.20 & 96.35 & 283.68 & 945.09 & \underline{17.35} \\
 & & + FastV~\cite{chen2024image} & 66.20 & 76.80 & 77.20 & 48.00 & 67.05 & \underline{265.11} & \underline{863.84} & 17.38 \\ 
 & & + Ours & \textbf{99.40} & \textbf{100.00} & \textbf{99.80} & \textbf{98.60} & \textbf{99.45} & \textbf{158.87} & \textbf{513.35} & \textbf{13.17} \\ \midrule
\multirow{4}{*}{X-VLA~\cite{zheng2026xvla}} 
 & \multirow{4}{*}{0.9B} & Baseline & \textbf{99.60} & \textbf{100.00} & \underline{98.00} & \underline{96.40} & \underline{98.50} & 151.32 & 387.52 & 10.73 \\
 & & + SparseVLM~\cite{zhang2025sparsevlm} & 99.20 & 82.00 & 97.00 & 90.60 & 92.20 & \underline{147.74} & 389.60 & \underline{10.20} \\
 & & + FastV~\cite{chen2024image} & 98.40 & \underline{99.60} & \underline{98.00} & 81.20 & 94.30 & 148.00 & \underline{385.21} & 10.31 \\ 
 & & + Ours & \underline{99.40} & \textbf{100.00} & \textbf{98.80} & \textbf{97.40} & \textbf{98.90} & \textbf{64.90} & \textbf{173.09} & \textbf{3.51} \\ \bottomrule
\end{tabular}
\vspace{-10pt}
\end{table*}

\subsection{Training-free \& Efficient MLP Importance Assessment}\label{subsec:mlp_importance}

    Applying a uniform pruning ratio to the MLP blocks across all layers is suboptimal, as individual layers exhibit varying degrees of parameter sensitivity and information density. 
Furthermore, computing gradient-based layer importance is often prohibitively expensive for edge \ymyi{devices} due to \ymyi{their} limited computational budgets and data privacy constraints.

To address this, \hesh{\SYS} evaluates \ymyi{layer} importance using only forward passes without access to training data, employing a metric inspired by the Effective Rank~\cite{roy2007effective} of MLP activations.
Our approach is grounded in the findings of RankMe~\cite{garrido2023rankme}, which demonstrate that the Effective Rank of neural representations is a robust indicator of downstream performance. This suggests that it can serve as a reliable proxy for a layer's expressivity; a higher rank implies that the representations are more diverse and capture essential features, whereas a lower rank indicates high redundancy. Specifically, for an activation matrix $A \in \mathbb{R}^{N \times M}$, the Effective Rank is defined using the Shannon entropy ($H_1$) of its singular value distribution:
\begin{equation}
\begin{aligned}
    \text{Effective Rank}(A) &= \exp(H_1(P)) \\
    &= \exp \left (-\sum_{k=1}^{\min(N,M)} p_k \ln p_k \right )
\end{aligned}
\end{equation}
where $p_k$ represents the normalized $k$-th singular value $\sigma_k$:
\begin{equation}
    p_k = \frac{\sigma_k(A)}{\sum_{m=1}^{\min(N,M)} \sigma_m(A)}
\end{equation}
Here, $\sigma_k(A)$ denotes the $k$-th singular value of $A$.
This formulation accurately captures the \textit{``spread''} of the singular value spectrum. By preserving high-rank blocks for representational diversity and aggressively pruning redundant low-rank ones, we can optimize the latency--success rate trade-off.

However, computing exact singular values via Singular Value Decomposition (SVD) introduces prohibitive latency on edge devices due to iterative numerical operations and GPU-CPU synchronization overhead. To bypass SVD, \SYS{} efficiently estimates the representational spread using the Participation Ratio, which is a computationally tractable variant of Effective Rank. Specifically, we leverage the symmetric Gram matrix $G=A^T A$, whose eigenvalues $\lambda_m$ directly correspond to the squared singular values of the original activation matrix $A$.

By defining the normalized eigenvalue distribution of the Gram matrix $G$ as $\tilde{p}_m = \lambda_m / \sum_n \lambda_n$ and replacing the Shannon entropy ($H_1$) with the Collision entropy ($H_2$), we derive the Participation Ratio (PR):
\begin{align}
    \text{PR} &= \exp(H_2(\tilde{P})) \nonumber  \\
    &= \exp ( -\ln ( \sum_m \tilde{p}_m^2 ) ) \nonumber \\
    &= \frac{1}{\sum_m \tilde{p}_m^2} 
    = \frac{\left(\sum_n \lambda_n\right)^2}{\sum_m \lambda_m^2}
\end{align}
Since the sum of eigenvalues equals the trace ($\sum_n \lambda_n = \text{tr}(G)$) and the sum of squared eigenvalues for a symmetric matrix $G$ equals the squared Frobenius norm ($\sum_m \lambda_m^2 = \|G\|_F^2$), the Participation Ratio (PR) of the $l$-th layer can be computed efficiently without any decomposition:
\begin{equation}
    \text{PR}_l = \frac{\left(\sum_n \lambda_n\right)^2}{\sum_m \lambda_m^2} = \frac{\{\text{tr}(G_l)\}^2}{\|G_l\|_F^2}
\end{equation}
where $G_l$ denotes the Gram matrix from the $l$-th layer. We normalize these Participation Ratios to obtain a relative importance score $\delta_l \in [0, 1]$ for the $l$-th layer:
\begin{equation}
    \delta_l = \frac{\text{PR}_l - \min_j(\text{PR}_j)}{\max_j(\text{PR}_j) - \min_j(\text{PR}_j)}
\end{equation}
Based on $\delta_l$, highly important MLP blocks ($\delta_l \approx 1$) are pruned conservatively to preserve capacity, while less important ones ($\delta_l \approx 0$) are pruned aggressively. 
\subsection{MLP Channel Reordering}\label{subsec:mlp_reordering}

Table~\ref{tab:latency} shows that VLM computational bottlenecks stem primarily from the MLP blocks.
\hesh{Given that these bottlenecks arise from the massive intermediate channels,
we adopt structured pruning, which is more \ymyi{efficient and} implementation-friendly than unstructured alternatives.}
\ymyi{Since access to training data is often restricted after deployment, offline calibration for pruning becomes infeasible.
To address this, 
\hesh{\SYS{}} dynamically reorders intermediate channels based on their channel importance metrics during inference, without requiring any calibration data.}
Flow-matching-based VLAs typically employ either Gated MLP~\cite{black2024pi0visionlanguageactionflowmodel,intelligence2025pi05visionlanguageactionmodelopenworld,shukor2025smolvla}
or 2-Layer MLP~\cite{zheng2026xvla}
with an activation function $\phi$:
\begin{align}
    \text{Gated MLP}(x) &= \text{Down}(\phi(\text{Gate}(x)) \odot \text{Up}(x)) \\
    \text{2-Layer MLP}(x) &= \text{fc2}(\phi(\text{fc1}(x)))
\end{align}
Unlike CATS~\cite{lee2024cats}, which assesses importance \hesh{based solely on} activation magnitude, \hesh{\SYS} accounts for the scaling effect of the subsequent projection layer. We define the importance $I_i$ of the $i$-th channel as the product of its activation magnitude and the corresponding projection weight norm:
\begin{align}
    I_i^{\text{gated}} &= \| \left[ \phi(\text{Gate}(x)) \odot \text{Up}(x) \right]_i \|_2 \cdot \| \mathbf{w}_{\text{Down}, i} \|_2 \\
    I_i^{\text{2-Layer}} &= \| \left[ \phi(\text{fc1}(x)) \right]_i \|_2 \cdot \| \mathbf{w}_{\text{fc2}, i} \|_2
\end{align}
where $[\cdot]_i$ denotes the $i$-th channel's activation tensor across the sequence length, and $\mathbf{w}_i$ \hesh{denotes the weight vector associated with the $i$-th intermediate channel.}
We sort the intermediate channels by importance and retain only the top fraction to ensure that critical features are preserved, thereby preventing performance degradation during the pruning process. While reordering could incur overhead, it is performed only during the initial forward pass or upon reaching the maximum number of allowed iterations, indicating a significant context shift.


\section{EXPERIMENTS}
\subsection{Experimental Setup}\label{exp:setup}
\ymyi{\textbf{Models:}} 
To validate the effectiveness of \hesh{\SYS{}}, we evaluated it on two representative state-of-the-art flow-matching-based VLAs: $\pi_{0.5}$~\cite{intelligence2025pi05visionlanguageactionmodelopenworld} and X-VLA~\cite{zheng2026xvla} with a default step size of $s=0.1$ (10 steps). 

\textbf{Simulation Benchmarks:} We employed the LIBERO benchmark~\cite{liu2023libero}, which comprises the Spatial, Object, Goal, and Long task suites, to evaluate diverse semantic and spatial reasoning capabilities. 
We set the curvature threshold $C_\text{th}$ to 0.15 for $\pi_{0.5}$ and 0.002 for X-VLA. 
These values were \ymyi{determined} via a simple grid search
\ymyif{, and our ablation studies demonstrate that performance is robust across a range of $C_\text{th}$ values. For a new model or benchmark, users can readily identify a near-optimal $C_\text{th}$ from a single profiling run by inspecting the resulting curvature distribution.}
All model checkpoints were obtained from the official LeRobot repository on Hugging Face~\cite{lerobot_pi05_libero_finetuned, lerobot_xvla_libero}.

\textbf{Hardware \& Metrics:} 
Following the evaluation protocol in~\cite{kim2025finetuning}, we evaluated each task over 50 rollouts.
To avoid prohibitively long evaluation times on the edge device (up to 12 hours per full benchmark), the average success rate was measured on a server-grade NVIDIA A10 GPU.
We additionally verified on a subset of tasks that the success rates on the edge-grade NVIDIA Jetson AGX Orin were comparable to those obtained on the A10. 
Latency was measured as the time required to generate a single action chunk on both the A10 GPU and the Jetson AGX Orin.
To reflect real-world edge constraints, the total system energy consumption (CPU, GPU, and motherboard) was measured on the Jetson AGX Orin. Both latency and energy consumption were averaged over 10 rollouts per task.

\subsection{Main Results on LIBERO Benchmark}
Table~\ref{tab:libero_main} presents the quantitative results on the LIBERO~\cite{liu2023libero} benchmark for $\pi_{0.5}$~\cite{intelligence2025pi05visionlanguageactionmodelopenworld} and X-VLA~\cite{zheng2026xvla}. Following our deployment setting, we compared against the original models and training-free acceleration methods that are directly applicable to pretrained checkpoints without additional retraining, distillation, or architecture-specific redesign, including SparseVLM~\cite{zhang2025sparsevlm} and FastV~\cite{chen2024image}.

On the Jetson AGX Orin, \hesh{\SYS{}} accelerates $\pi_{0.5}$ by $1.87\times$ (reducing latency from 958.54 ms to 513.35 ms) while slightly improving the average success rate to $99.45\%$ and effectively reducing energy consumption \ymyi{by $1.51\times$} from 19.83 J to 13.17 J. These efficiency gains are even more pronounced for smaller architectures like X-VLA,  
\hesh{which are typically latency-bound due to kernel launch and memory access overheads.}
Previous methods~\cite{zhang2025sparsevlm, chen2024image} target compute reduction solely within the VLM but \hesh{fail to mitigate the underlying execution overheads.} Consequently, they yield negligible speedups (e.g., FastV~\cite{chen2024image} reduces X-VLA latency by merely 2.3~ms on Jetson Orin). Conversely, by dynamically reducing flow matching steps, \hesh{\SYS{}} bypasses a substantial number of iterative forward passes, effectively \hesh{alleviating} the \hesh{kernel launch and memory access overheads.} For X-VLA, 
\hesh{this advantage of \SYS{} translates} to a $2.24\times$ latency reduction (from 387.52 ms to 173.09 ms) and a significant $3.06\times$ decrease in energy consumption (from 10.73 J to 3.51 J), while ensuring near-perfect success rates across all complex sub-tasks.

\camera{\ymyif{\SYS{} also} maintains stable inference \ymyif{timing}. On the Jetson AGX Orin, the standard deviations of per-step latency are 19.12~ms for $\pi_{0.5}$ and 21.06~ms for X-VLA, \ymyif{demonstrating} the \ymyif{timing} stability \ymyif{required} for real-world deployment.}

\begin{table}[t]
\centering
\setlength{\tabcolsep}{3pt}
{
\caption{Ablation study of \SYS{} with $\pi_{0.5}$ on the LIBERO benchmark using Jetson AGX Orin.}
\label{tab:ablation}
\begin{tabular}{@{}lcccc@{}}
\toprule
\textbf{Variant} &$C_\text{th}$ &\textbf{Avg SR (\%)} & \textbf{Latency (ms)} & \textbf{Energy (J)} \\ \midrule
Baseline (10-step) & - & 98.65 & 958.54 & 19.83 \\
\midrule
\textbf{Ours (+PR, +Adp)} & \multirow{3}{*}{0.15} & 99.45 & 513.35 & 13.17 \\
Ours (+SVD, +Adp) &  & 99.65 & 894.78 & 23.60 \\
Ours (+Fxd, +Adp) &  & 99.35 & 492.00 & 12.23 \\ \midrule
Ours (-Pruning, +Adp)& \multirow{2}{*}{0.15} & 98.65 & 532.32 & 13.33 \\
Ours (+PR, -Adp)&  & 99.25 & 925.71 & 19.10 \\ \midrule
Ours (+PR, +Adp)& 0.20 & 99.20 & 428.24 & 10.48 \\ 
Ours (+PR, +Adp)& 0.25 & 98.40 & 413.26 & 9.96 \\ \midrule
Baseline (1-step) & - &  97.05 & 447.45 & 11.43 \\
Baseline (5-step) & - & 98.00 & 677.56 & 15.31 \\
Baseline (Random-step) & - & 98.65 & 694.13 & 15.49 \\ \bottomrule
\end{tabular}
}
\vspace{-15pt}
\end{table}

\subsection{Ablation Study}\label{exp:ablation}
To evaluate the specific impact of each component, we conducted ablation studies on $\pi_{0.5}$ using the Jetson AGX Orin. The results are summarized in Table~\ref{tab:ablation}.

\textbf{Impact of Efficient Assessment:} While SVD \ymyi{better} identifies critical layers leading to a high success rate (99.65\% SR), 
\ymyi{it offers limited latency improvements compared to our Participation Ratio-based pruning method (\ymyi{894.78}~ms vs 513.35~ms)}.
\hesh{It also} incurs an unacceptable Time-to-First-Action (TTFA) of 4939.32~ms and 23.60~J energy cost due to decomposition \ymyi{overheads}. 
\ymyi{In contrast, the Participation Ratio (PR) approach obviates the need for explicit decomposition, utilizing computationally efficient trace and norm calculations instead.}
This yields a 5.63$\times$ TTFA speedup (877.13~ms) and lowers energy usage to 13.17~J without meaningful performance degradation. Although PR adds a marginal 21~ms latency overhead compared to the fixed variant (+Fxd), it effectively preserves a 99.45\% SR by accurately prioritizing representational density.

\textbf{Necessity of Adaptive Inference and Pruning:} Our adaptive components jointly optimize throughput and reliability. In particular, PR-based pruning safeguards model robustness. Removing the pruning component (-Pruning, +Adp) causes the success rate to revert to the 98.65\% baseline, alongside slight increases in latency (532.32~ms) and energy (13.33~J). Simultaneously, adaptive inference is primarily responsible for latency control. Excluding this \ymyi{(i.e., +PR, -Adp)} results in a substantial latency increase to 925.71~ms and an energy cost of 19.10~J, nearly matching baseline levels.
Although the success rate remains at 99.25\%, these observations suggest that adaptive inference is \ymyi{a key enabler for achieving real-time performance on edge devices}, as its removal largely \ymyi{nullifies} the latency gains of \ymyi{the proposed} framework.

\begin{figure}[t]
    \centering
    \begin{subfigure}{\linewidth}
        \centering
        \includegraphics[width=0.8\linewidth]{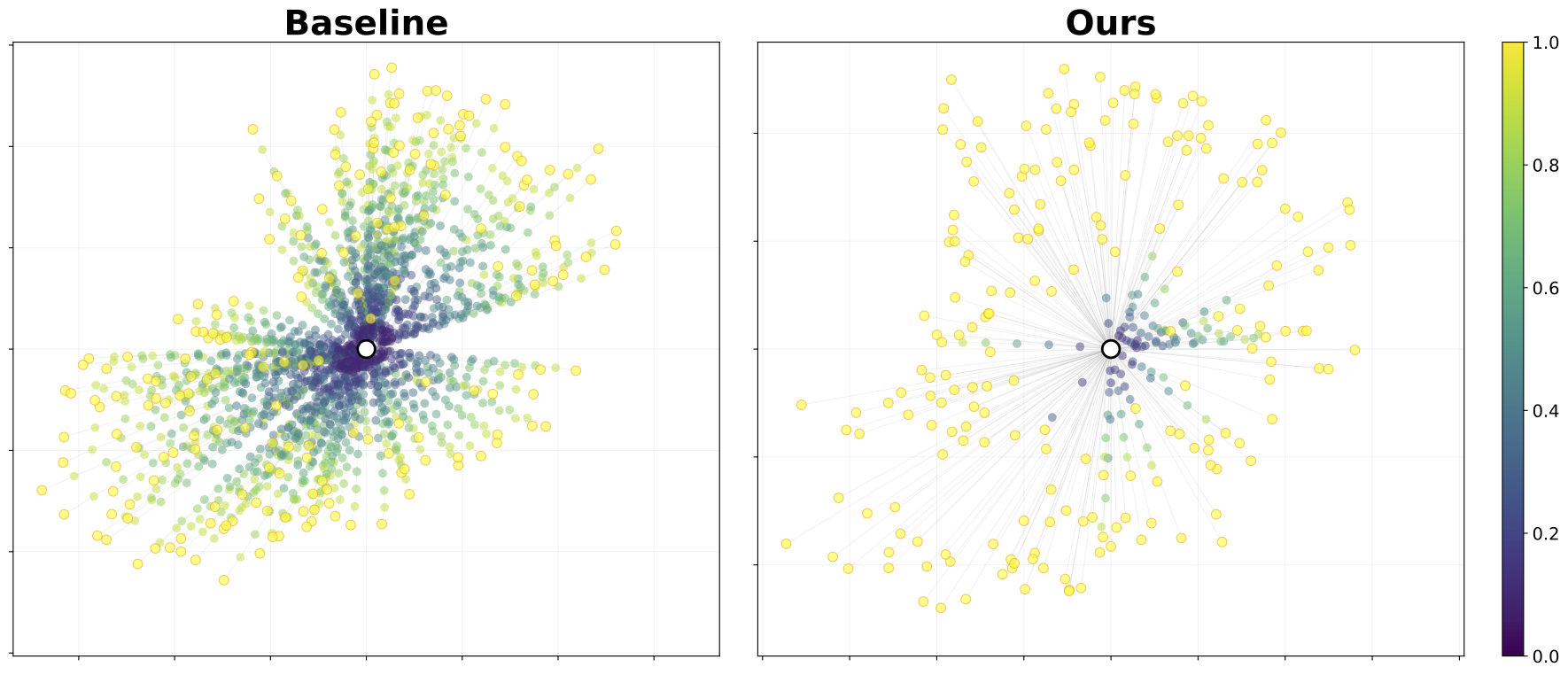}
        \caption{Trajectories of $\pi_{0.5}$ ($C_\text{th}=0.15$)}
        \vspace{5pt}
        \label{fig:sub_top}
    \end{subfigure}
    \begin{subfigure}{\linewidth}
        \centering
        \includegraphics[width=0.8\linewidth]{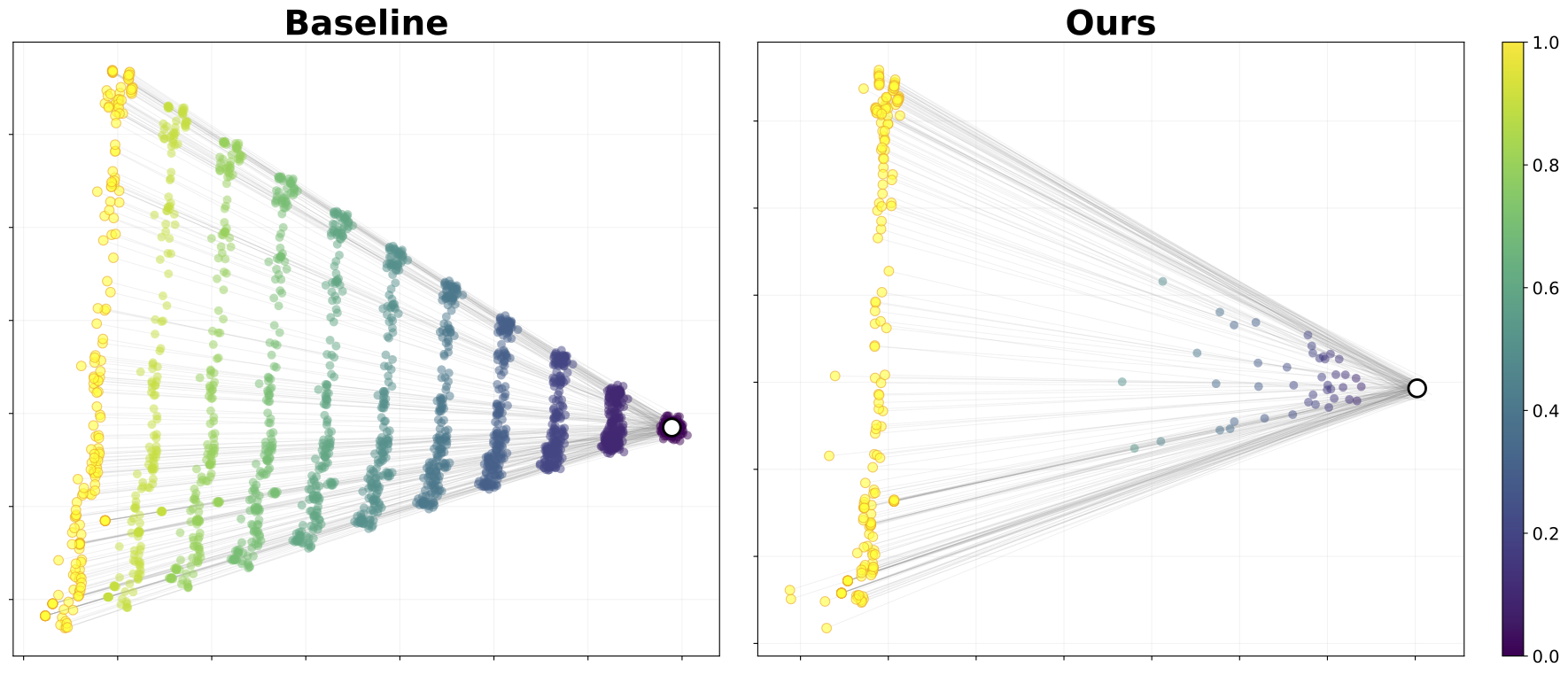}
        \caption{Trajectories of X-VLA ($C_\text{th}=0.002$)}
        \label{fig:sub_bottom}
    \end{subfigure}
    \caption{Visualization of trajectories for $\pi_{0.5}$ and X-VLA during LIBERO-Long execution with default step size $s=0.1$.
    The color gradient from dark to light represents the flow matching time variable $\tau \in [0, 1]$.}
    \vspace{-20pt}
    \label{fig:fm_analysis}
\end{figure}

\textbf{Sensitivity of Curvature Threshold:} The curvature threshold $C_{\text{th}}$ serves as a predictable control knob for the latency--success rate trade-off. 
Increasing $C_{\text{th}}$ from 0.15 to 0.25 further reduces latency to 413.26~ms, at the cost of a marginal decline in the success rate to 98.40\%. 
This stability suggests that the framework is not sensitive to threshold selection, providing flexible optimization across different hardware constraints without excessive \hesh{dependence} on the specific operating point. Depending on deployment requirements, users may prioritize lower latency or higher success rate; \SYS{} provides this flexibility through $C_{\text{th}}$, which can be selected via a lightweight grid search.

\textbf{Comparison with Static Step Reduction:} To determine if simply reducing flow matching steps can match our adaptive approach, we evaluated baseline variants with 1, 5, and random (1–10) flow matching steps. 
Results in Table~\ref{tab:ablation} indicate that static reduction leads to decreased task reliability. For instance, a 1-step baseline achieves 447.45~ms latency, but its 97.05\% success rate is \camera{lower than} that of our adaptive method (99.45\%). \camera{With 5-step or random-step configurations, the baseline success rates become} \camera{comparable to that of our adaptive method, but their latencies increase to 677.56 ms and 694.13 ms, respectively.}
These observations \camera{suggest} that naive step reduction may limit the model's capacity to navigate complex \hesh{trajectories}. Conversely, \SYS{} concentrates \ymyi{computational} resources on critical segments, \camera{improving} the latency--success rate trade-off.


\subsection{Real-World Robot Experiments}

To bridge the gap between simulation and physical deployment, we conducted real-world validation using the SO-ARM101 robot~\cite{knight2024standardopen}. The entire inference process was executed on the NVIDIA Jetson AGX Orin, serving as a representative edge computing platform for mobile manipulators.
We employed SmolVLA~\cite{shukor2025smolvla} as the base policy and fine-tuned it via imitation learning using 50 teleoperated expert demonstrations per task \hesh{on a single A10 GPU}. To evaluate the model's versatility, we designed four distinct manipulation tasks requiring precise coordination:
\begin{itemize}[leftmargin=*]
    \item \textit{Task 1: Pick up the banana and place it in the basket.}
    \item \textit{Task 2: Grab the socks and put them into the laundry bin.}
    \item \textit{Task 3: Pick up the sausage and place it on the plate.}
    \item \textit{Task 4: Pick up the cup and place it on the plate.}
\end{itemize}

For the quantitative comparison, each task was evaluated over 30 independent trials. We measured the success rate and average latency, defined as the time required to generate a single action chunk on the Jetson AGX Orin. A uniform curvature threshold of $0.075$ was applied across all tasks.

\begin{figure}[t]
    \centering
    \includegraphics[width=0.9\columnwidth]{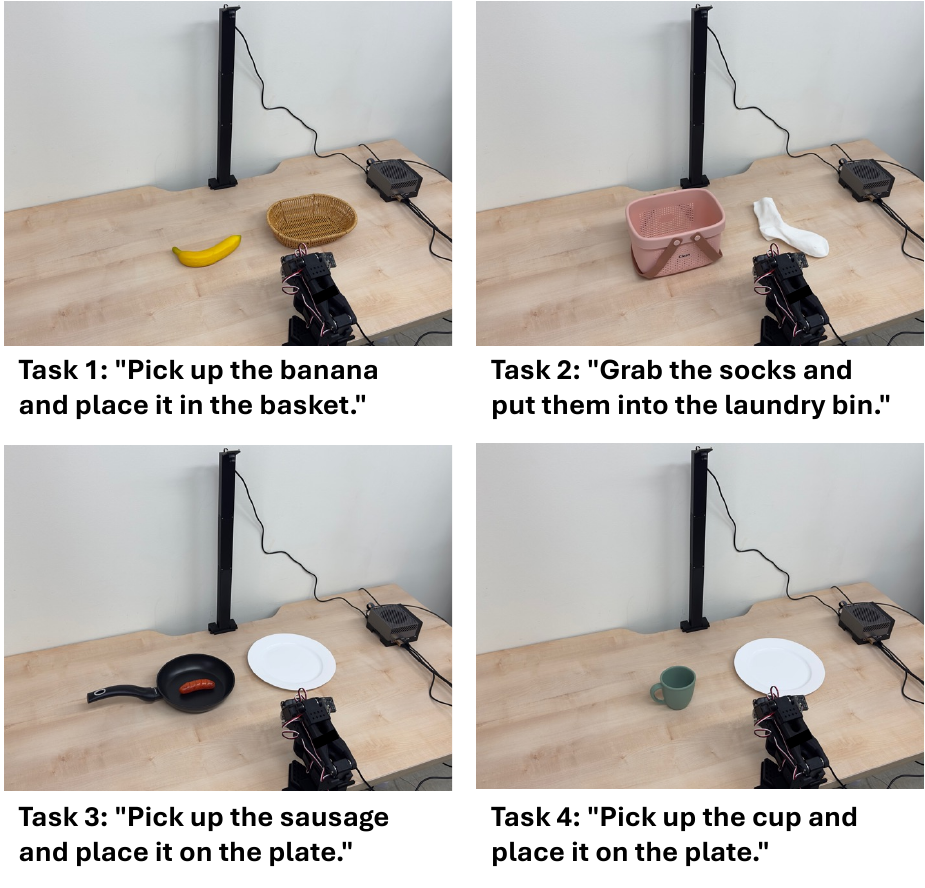} 
    \caption{Task description of real robot experiments}
    \label{fig:robot_settings}
    \vspace{-20pt}
\end{figure}

As summarized in Table~\ref{tab:real_world}, our framework achieves a substantial reduction in \ymyi{latency}, averaging 488.49 ms compared to the 806.73 ms of the baseline, which translates to a 1.65$\times$ speedup. 
Notably, this acceleration is achieved while also improving the overall average success rate from $87.50\%$ to $90.00\%$, effectively offsetting the marginal performance degradation observed in specific scenarios (Tasks 3 and 4). 
\camera{This degradation primarily stems from coarser action trajectories induced by the larger integration steps, and can be readily mitigated by lightweight post-processing techniques such as action smoothing, which incur negligible computational cost.}
These results underscore the efficacy of our adaptive approach, demonstrating that it can significantly accelerate inference speed without compromising overall operational reliability in real-world robotic deployments.

\subsection{Analysis of Adaptive Inference}\label{exp:adp_anlys}

We visualized the actual \hesh{trajectories} generated by \ymyi{flow-matching-based} VLAs to analyze the behavior of \SYS. Fig.~\ref{fig:fm_analysis} illustrates 250 randomly sampled \hesh{trajectories} recorded during the LIBERO-Long benchmark~\cite{liu2023libero} for both $\pi_{0.5}$~\cite{intelligence2025pi05visionlanguageactionmodelopenworld} and X-VLA~\cite{zheng2026xvla} under the same setup described in Sec.~\ref{exp:setup}. 

In the \ymyi{baseline} (left), both models consistently execute a high number of flow matching iterations, leading to a substantial number of forward passes in the Action Expert. In contrast, our method (right) generates clean actions with significantly fewer forward passes while adaptively allocating additional steps only when necessary. 

\begin{table}[t]
\centering
\caption{Real-world experimental results on the SO-ARM101 robot using SmolVLA evaluated on the Jetson AGX Orin. Success rate is reported as (Successes / Trials).}
\label{tab:real_world}
\begin{tabular}{@{}lcccc@{}}
\toprule
\multirow{2}{*}{\textbf{Task}} & \multicolumn{2}{c}{\textbf{Success Rate}} & \multicolumn{2}{c}{\textbf{Latency (ms)}} \\ \cmidrule(lr){2-3} \cmidrule(l){4-5} 
 & Baseline & \textbf{Ours} & Baseline & \textbf{Ours} \\ \midrule
Task 1 & 25/30 (83.3\%) & \textbf{29/30 (96.7\%)} & 803.49 & \textbf{496.79} \\
Task 2 & 26/30 (86.7\%) & \textbf{28/30 (93.3\%)} & 811.53 & \textbf{447.07} \\
Task 3 & \textbf{29/30 (96.7\%)} & 27/30 (90.0\%) & 805.31 & \textbf{476.00} \\
Task 4 & \textbf{25/30 (83.3\%)} & 24/30 (80.0\%) & 806.58 & \textbf{534.10} \\ \midrule
\textbf{Average} & 87.50\% & \textbf{90.00\%} & 806.73 & \textbf{488.49} \\ \bottomrule
\end{tabular}
\end{table}

\begin{figure}[t]
    \centering
    \begin{subfigure}{0.3\columnwidth}
        \centering
        \includegraphics[width=\linewidth]{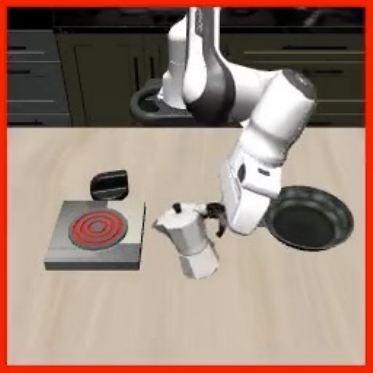}
        \caption{Grab Object}
        \label{fig:sub_grab}
    \end{subfigure}
    \hfill
    \begin{subfigure}{0.3\columnwidth}
        \centering
        \includegraphics[width=\linewidth]{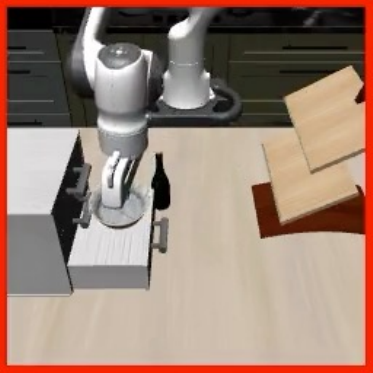}
        \caption{Place Object}
        \label{fig:sub_place}
    \end{subfigure}
    \hfill
    \begin{subfigure}{0.3\columnwidth}
        \centering
        \includegraphics[width=\linewidth]{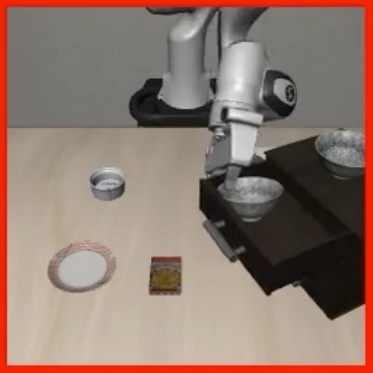}
        \caption{Failure}
        \label{fig:sub_fail}
    \end{subfigure}
    \caption{Qualitative analysis of critical task segments where our framework adaptively allocates maximum inference steps}
    \vspace{-17.5pt}
    \label{fig:case_study}
\end{figure}

Furthermore, we analyzed the specific task segments where these additional steps are triggered, \hesh{particularly} focusing on instances where the model reaches the maximum step limit. Our findings reveal that the model intensifies its computational effort during critical task moments, such as grasping~(\ref{fig:sub_grab}) and placing~(\ref{fig:sub_place}) objects. Moreover, additional steps were dynamically invoked when the environment shifted unexpectedly—for instance, when an object fell during a pick-up attempt~(\ref{fig:sub_fail}). These results demonstrate that the curvature of the \hesh{trajectories} can serve as an intuitive proxy for perceived difficulty and action confidence, reducing reliance on complex, explicitly modeled difficulty estimators.


\section{CONCLUSION}
We presented \hesh{\SYS}, a training-free, online adaptive framework for \ymyi{fast yet accurate flow-matching-based} VLAs. 
Inspired by human cognition, our method autonomously regulates computational effort by utilizing \hesh{flow-matching trajectory curvature} to adjust inference steps and MLP pruning ratios.
On the Jetson AGX Orin, we achieved $1.87\times$ to $2.24\times$ 
speedup with negligible success rate degradation on the LIBERO benchmark and $1.65\times$ 
speedup in real-world experiments. By eliminating the need to access training data for fine-tuning or calibration, \SYS{} offers a highly practical solution for deploying \ymyi{flow-matching-based} VLAs on resource-constrained robotic systems.


\section*{ACKNOWLEDGMENT}

In accordance with the IROS guidelines on Generative AI usage, we explicitly state that we utilized Gemini 3 to refine the grammatical structure, improve readability, and generate specific illustrative elements for Fig.~\ref{fig:paper_concept} and Fig.~\ref{fig:method_overview}.
This work was supported by Institute of Information \& Communications Technology Planning \& Evaluation (IITP) grant funded by the Korea government (MSIT) (No. RS-2025-02263167, and No. IITP-RS-2026-25547954 under the Artificial Intelligence Innovation Human Resources Development).


\bibliographystyle{IEEEtran}
\bibliography{ref}

\end{document}